\documentclass[11pt,letterpaper]{article}
\usepackage[top=0.85in,left=1.6in,footskip=0.5in,marginparwidth=1.1in]{geometry}
\pdfoutput=1

\usepackage[utf8]{inputenc}

\usepackage{cite}
\usepackage{makecell}
\usepackage{multirow}
\usepackage[capitalise]{cleveref}
\usepackage{CJKutf8}
\usepackage{xcolor}

\usepackage{nameref,hyperref}

\usepackage[right]{lineno}

\usepackage{microtype}
\DisableLigatures[f]{encoding = *, family = * }

\usepackage{changepage}

\usepackage[aboveskip=1pt,labelfont=bf,labelsep=period,singlelinecheck=off]{caption}

\makeatletter
\renewcommand{\@biblabel}[1]{\quad#1.}
\makeatother

\usepackage{lastpage,fancyhdr,graphicx}
\usepackage{epstopdf}
\fancyheadoffset[L]{2.25in}
\fancyfootoffset[L]{2.25in}

\usepackage{color}

\definecolor{Gray}{gray}{.25}

\usepackage{graphicx}

\usepackage{sidecap}

\usepackage{wrapfig}
\usepackage[pscoord]{eso-pic}
\usepackage[fulladjust]{marginnote}
\reversemarginpar

\usepackage{amsmath}
\usepackage{url}

\newcommand{\NA}{---}

\begin{document}
\begin{CJK*}{UTF8}{gbsn}
\vspace*{0.35in}

\begin{flushleft}
{\Large
\textbf\newline{PADME: A Deep Learning-based Framework for Drug-Target Interaction Prediction}
}
\newline
\\
Qingyuan Feng\textsuperscript{1},
Evgenia Dueva\textsuperscript{2},
Artem Cherkasov\textsuperscript{2,3,*},
Martin Ester\textsuperscript{1,2,*}
\\
\bigskip
\bf{1} School of Computing Science, Simon Fraser University, 8888 University Drive, Burnaby, BC, V5A1S6, Canada 
\\
\bf{2} Vancouver Prostate Centre, Vancouver, BC V6H3Z6, Canada
\\
\bf{3} Department of Urologic Sciences, University of British Columbia, Vancouver, BC V5Z1M9, Canada
\\
\bigskip
* ester@sfu.ca; artc@interchange.ubc.ca

\end{flushleft}

\section*{Abstract}
\emph{In silico} drug-target interaction (DTI) prediction is an important and challenging problem in biomedical research with a huge potential benefit to the pharmaceutical industry and patients. Most existing methods for DTI prediction including deep learning models generally have binary endpoints, which could be an oversimplification of the problem, and those methods are typically unable to handle cold-target problems, i.e., problems involving target protein that never appeared in the training set. Towards this, we contrived PADME (Protein And Drug Molecule interaction prEdiction), a framework based on Deep Neural Networks, to predict real-valued interaction strength between compounds and proteins without requiring feature engineering. PADME takes both compound and protein information as inputs, so it is capable of solving cold-target (and cold-drug) problems. To our knowledge, we are the first to combine Molecular Graph Convolution (MGC) for compound featurization with protein descriptors for DTI prediction. We used multiple cross-validation split schemes and evaluation metrics to measure the performance of PADME on multiple datasets, including the ToxCast dataset, and PADME consistently dominates baseline methods. The results of a case study, which predicts the binding affinity between various compounds and androgen receptor (AR), suggest PADME's potential in drug development. The scalability of PADME is another advantage in the age of Big Data.

\linenumbers
\nolinenumbers

\section{Introduction}

Finding out the interaction strengths between compounds (candidate drugs) and target proteins is of crucial importance in the drug development process. However, it is both expensive and time-consuming to be done in wet lab experiments, while virtual screening using computational (also called \emph{``in silico''}) methods to predict the interactions between compounds and target proteins can greatly accelerate the drug development process at a significantly reduced cost. Indeed, machine learning models for drug-target interaction (DTI) prediction are often used in computer-aided drug design \cite{Ding2014}.

Datasets used for training and evaluating machine learning models for DTI prediction often include compounds' interaction strengths with enzymes, ion channels, nuclear receptors, etc \cite{Yamanishi2008}. Traditionally, these datasets contain binary labels for the interaction of certain drug-target pairs, with 1 indicating a known interaction. Recently, the community has also explored the usage of datasets with real-valued interaction strength measurements \cite{Pahikkala2014, He2017}, which include the Davis dataset \cite{Davis2011} that uses the inhibition constant (\(K_i\)), the Metz dataset \cite{metz2011} that uses the dissociation constant (\(K_d\)) and the KIBA dataset \cite{tang2014} whose authors devised their own measurement index.

Existing traditional machine learning methods for predicting DTI can be roughly divided into similarity-based and feature-based approaches, and most of them formulate the problem as a classification problem. Similarity-based methods depend on the assumption that compounds with similar structures should have similar effects. Feature-based methods construct feature vectors as input, which are generated by combining descriptors of compounds with descriptors of targets, and the feature vectors serve as inputs for algorithms such as support vector machine (SVM) \cite{He2017}. 

SimBoost \cite{He2017} and KronRLS \cite{Pahikkala2014} are two state-of-the-art methods for DTI prediction. Both of them have single outputs. KronRLS is based on Regularized Least Squares and utilizes the similarity matrices for drugs and targets to get the parameter values. SimBoost is a feature-based method, but in its feature construction, similarity matrices of the drugs and those of targets are also involved. These methods can both predict continuous values and binarized values. However, these methods either simply rely on similarities, or require expert knowledge to define the relevant features of proteins and compounds, called ``feature engineering''. Additionally, they are often unable to model highly complex interactions within compound molecules \cite{mayr2016} and between the compounds and their target proteins. 

Deep Neural Networks (DNN) promise to address these challenges.

Deep learning, the machine learning method based on DNN, has been enjoying ever-rising popularity in the past few years. It has seen wide and exciting applications in computer vision, speech recognition, natural language processing, reinforcement learning, and drug-target interaction prediction. DNNs can automatically extract important features from the input data, synthesize and integrate low-level features into high-level features, and capture complicated nonlinear relationships in a dataset \cite{lecun2015, schmidhuber2015}. Deep learning-based DTI prediction has been shown to consistently outperform the existing methods and has become the new ``golden standard'' \cite{dahl2014, unterthiner2014, ma2015}. 

The current deep learning approaches to drug-target interaction prediction can be roughly categorized based on their neural network types and prediction endpoints. Simple feedforward neural networks, Convolutional Neural Networks (CNN) and Recurrent Neural Networks (RNN) have been adopted in various papers \cite{xu2017, lenselink2017}. To our knowledge, almost all existing deep learning methods, except those that have 3D structural information as input, treat the problem as a classification problem, most of which are binary, namely active/inactive. Though there are deep learning models using 3D structural information that yield good results in regression problems \cite{wallach2015, gomes2017}, the requirement of 3D structural information limits the applicability of a model since such information is not always available, so we do not consider them in this paper. As deep learning for DTI is still in its infancy, the current models have several disadvantages.

\textbf{First}, formulating the problem as a classification problem has some inadequacies: obviously, the classification result depends on a predefined binarization threshold, which introduces some arbitrariness into the data; some useful information is lost, for instance, true-negative and missing values may not be discriminated in some chemical datasets  \cite{Pahikkala2014, He2017}. On the other hand, if we formulate it as a regression problem, not only can we avoid the problems above, but given the regression results, the real-valued outputs can be easily converted to produce a ranking or classification. Some existing non-DNN methods formulate the problem as a regression problem, in which the interaction strength between the drug molecule and the target protein is a real number, serving as the regression target \cite{He2017}. Common real-valued interaction strength metrics include \(K_i\),\(K_d\), etc.

\textbf{The second problem} is that most of the existing deep learning methods do not incorporate the target protein information into the network, except very few recent works, like Wen et al. \cite{wen2017} and Lenselink et al. \cite{lenselink2017}. As a result, the models are unable to solve the ``cold target'' problem, i.e. to predict the drug-target interactions for target proteins absent in the training dataset. In the field of proteochemometrics, this idea of combining protein and compound information as the input to the model was widely used, though most of them do not use DNN models \cite{cortes2015, wikberg2004, van2011, qiu2016}.

A recent model, DeepDTI \cite{wen2017}, addressed the second problem by combining the protein information with the compound feature vector. It uses the classical Extended-Connectivity Fingerprint (ECFP) \cite{rogers2010} for describing compounds, which relies on a fixed hashing function and cannot adjust to specific problems at hand. DeepDTI concatenates ECFP and Protein Sequence Composition (PSC) descriptors \cite{cao2013} (describing the target proteins' sequence information) to construct a feature vector, which is fed into a Deep Belief Network (DBN) to predict a binary endpoint. DeepDTI outperformed the state-of-the-art methods on a dataset extracted from DrugBank. Another slightly later work used feedforward networks with the combined compound and protein feature vector as the input and binary classification result as the output \cite{lenselink2017}. Their compound and protein feature vectors also required lots of feature engineering and expert knowledge.

In this paper, we propose PADME (Protein And Drug Molecule interaction prEdiction), a deep learning-based framework for predicting DTI, which can be roughly categorized into the feature-based methods. PADME overcomes the limitations of the existing methods by predicting real-valued interaction strengths instead of binary class labels, and, to address the cold-start problems (drugs or targets that are absent from the training set but appear in the test set), PADME utilizes a combination of drug and target protein features/fingerprints as the input vector, where no feature engineering is required. The drug and target vectors can be generated from SMILES representation and Amino Acid sequence, respectively, without loss of information. Unlike DeepDTI which uses DBN, PADME uses a feedforward network, mainly composed of ReLU layers, to connect the input vector to the output layer. PADME adopts Molecular Graph Convolution (MGC) which is more flexible than ECFP, because it learns the mapping function from molecular graph representations to feature vectors \cite{Duvenaud2015, kearnes2016, altae2017} . Similar to DeepDTI, we used Protein Sequence Composition (PSC) descriptor to represent the protein. To the best of our knowledge, this work is the first to integrate MGC with protein descriptors for the DTI problem. In addition to the kinase inhibitor datasets used by previous researchers, we also used the ToxCast dataset \cite{EPA2015}, which has a much larger variety of proteins and could be another useful benchmarking dataset for future researches of the same type. 

We conducted computational experiments with multiple cross-validation settings and evaluation metrics. The results demonstrated the superiority of PADME over baseline methods across all experimental settings. Besides, PADME is more scalable than SimBoost and KronRLS since it does not rely on computationally expensive similarity matrices and can accommodate multiple outputs.As a case study, we also applied PADME to predict the binding affinity between some compounds and the androgen receptor (AR). We examined the top compounds among them and confirmed this prediction through literature research, suggesting that the predictions of PADME have practical implications. 

The subsequent sections of the paper are organized as follows. The Method section will introduce the methods for compound featurization, protein featurization, and network structure. The Experiments section will present the experiments conducted, introducing the baseline methods, datasets used, experimental design, and the experimental results. The Discussion section clarifies some implementation and design choices, and outlines possible future directions to further this work. The last section concludes the whole paper.

\section{Method}

\begin{figure}[ht] 


\includegraphics[width=\textwidth]{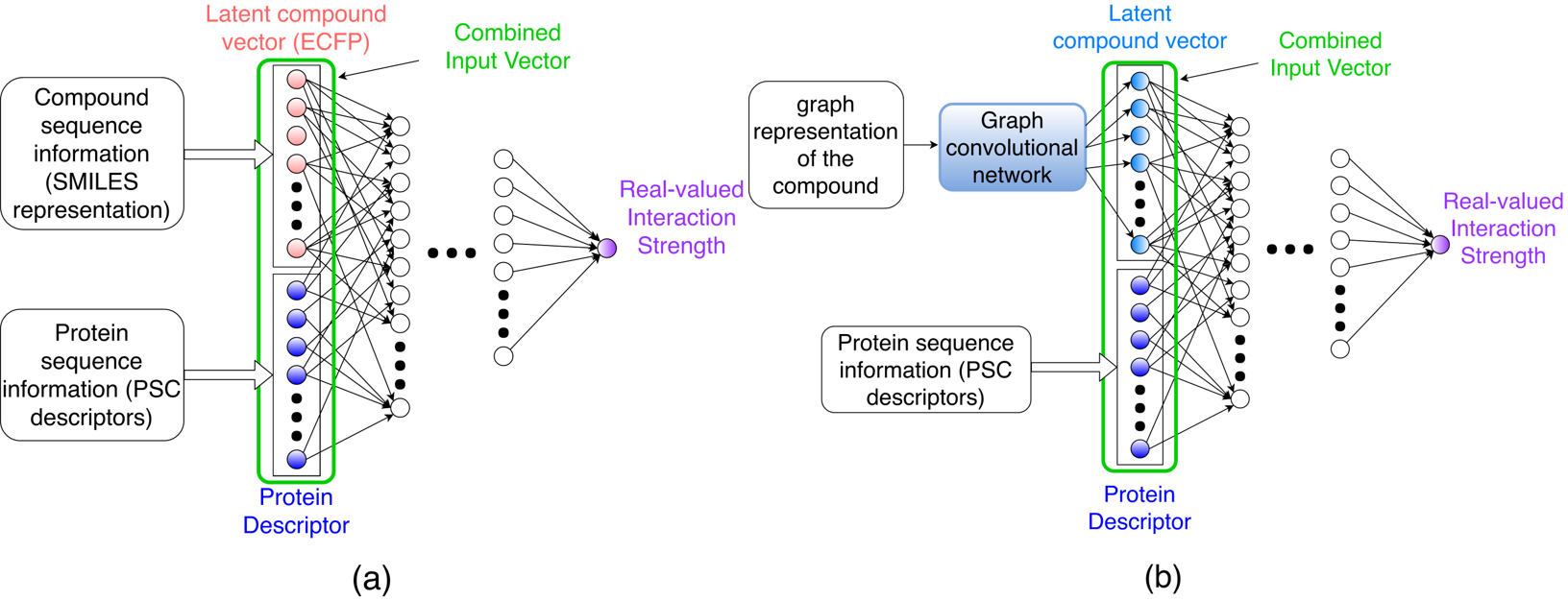}

\caption{\color{Gray}\textbf{a) PADME-ECFP architecture}. The Extended-Connectivity Fingerprint was used as the molecular input to the model. \textbf{b) PADME-GraphConv architecture.} Note that the graph convolutional network generating the latent molecular vector is trained together with the rest of the network, while the protein descriptor generation process is independent of the training of the network. The black dots represent omitted neurons and layers.}

\label{network} 

\end{figure}

PADME is a deep learning-based DTI prediction model which uses the combined small-molecule compound (candidate drug) and target protein feature vectors. We consider two variants of PADME with either Molecular Graph Convolution (MGC) or ECFP \cite{rogers2010} as the compound featurization method. For the protein, we use Protein Sequence Composition (PSC) descriptor \cite{cao2013}. In fact, PADME is compatible with all kinds of protein descriptors and molecular featurization methods. The compound vector is concatenated with a target protein vector to form the Combined Input Vector (CIV) for the neural network. PADME predicts a real-valued interaction strength, i.e., it solves a DTI regression problem. The structure of the network is shown in \autoref{network}: if we use the MGC network to get the molecular vector, that network will be trained together with the feedforward network connecting the CIV to the prediction endpoint in an end-to-end fashion. 

\subsection{Compound Featurization}

There has been a lot of research on representing small molecules (compounds) as a descriptor or fingerprint. Among the traditional molecular descriptors and fingerprints, ECFP \cite{rogers2010} is widely adopted as the state-of-the-art method for compound featurization \cite{Duvenaud2015}, and was also used in DeepDTI \cite{wen2017}. However, it has a fixed set of mapping and hashing functions, unable to be tailored for the specific task at hand.

DNN, especially MGC, can be used to generate more flexible feature vectors. Instead of depending only on the molecule, compound feature vectors generated using DNN depend on both the molecule and the prediction task (Boolean or continuous). DNN-derived feature vectors can outperform the ECFP baseline and at times offer some good interpretability \cite{Duvenaud2015, altae2017, xu2017}. 

MGC \cite{Duvenaud2015, kearnes2016, altae2017} is an extension of Convolutional Neural Network which learns a vector representing the compound from the graph-based representation of the molecule. In the graph representation of molecules, the atoms are denoted by nodes, while the bonds are denoted by edges. MGC takes into account the neighbors of a node when computing the intermediate feature vector for a specific node, and the same operation is applied to the neighborhood of each node (atom), hence it is analogous to ordinary convolutional networks typically used in Computer Vision \cite{Duvenaud2015, Goodfellow-et-al-2016}. Due to the GraphConv model \cite{wu2018} among MGC models being more recent and popular with an easier implementation, we use the GraphConv model as a representative of MGC under the time and resource constraints.

We applied both types of compound featurization methods: ECFP and GraphConv, and compared their performances. They can both be converted from SMILES representation without loss of information.

\subsection{Target Protein Featurization} 
There exist many schemes to represent the target protein as a feature vector based on its amino acid sequence information. DeepDTI \cite{wen2017} used Protein Sequence Composition (PSC) descriptor, which has 8420 entries for each protein, consisting of amino acid composition (AAC), dipeptide composition (DC), and tripeptide composition (TC) \cite{cao2013}. It captures rich information and does not transform the protein as much as some other protein descriptors (which implies less human knowledge required and less information loss), which we think could be a desirable attribute as the input to a neural network. In addition to the 8420 entries for each protein sequence, we added an additional binary entry signaling the phosphorylation status so that the Davis dataset (see below) can be represented more accurately, with `1' denoting phosphorylated, resulting in 8421 entries in total.

Mousavian et al. \cite{mousavian2016} used PSSM (Position Specific Scoring Matrix) descriptor to represent the protein, which focuses on dipeptide sequences and is related to the evolutionary history of proteins \cite{sharma2013}. It is observed that PSSM performed pretty well. Other popular protein sequence descriptors include Autocorrelation, CTD (Composition, Transition and Distribution) descriptor, Quasi-sequence order, etc \cite{cao2013}.

As PSC contains rich information (like tri-peptide sequence occurrence) with high dimensionality, and has already shown promising performance in deep learning-based models for DTI prediction \cite{wen2017}, we use PSC in this research. There could be future comparisons of the performance of PSC and other protein featurization methods as an extension to this work.

\subsection{Architecture of the Deep Neural Network} 

PADME uses a feedforward neural network taking the CIV as the input. The PADME architecture has one output neuron per prediction endpoint, i.e., one output neuron for most datasets, and 61 output neurons for the ToxCast dataset (see below). DNNs with single output neuron are called single-task networks, and those with multiple output neurons are called multi-task networks. Although we only consider the DTI regression problem in this paper, PADME can also be used for constructing classification models with minimal changes, either by binarizing the continuous prediction results or by directly using a softmax/sigmoid layer as the output layer, of which the latter could be more preferable.

For regularization, we use Early Stopping, Dropout and Batch Normalization techniques \cite{Goodfellow-et-al-2016}. Hyperparameters like dropout rates are automatically searched to find the best set of them before running cross-validation, as elaborated in the ``Experimental Design'' section. Adam optimizer \cite{kingma2014} was used to train the network. The activation functions used for fully connected layers are all Rectified Linear Units (ReLU).

\subsection{Time Complexity}
\label{subsec: tc}
PADME does not require drug-drug or target-target similarity matrices or matrix factorization, so it is much more scalable than KronRLS and SimBoost, to be introduced in the next section. Suppose there are \(n\) compounds and \(m\) proteins, since KronRLS and SimBoost need the similarity matrices, both of the methods have at least \(O(n^2 + m^2)\) time and space complexity, SimBoost involves matrix factorization so it is even more expensive. But in each epoch of PADME's training process, the time complexity only depends on the number of drug-target pairs in the training set, which, in the best case, is \(O(max(n,m))\), and \(O(nm)\) in the worst case. There is no closed-form relationship between the number of epochs until convergence and \(n\) or \(m\), so the number of epochs cannot be analyzed theoretically. However, we observe that, in practice, this number is possibly sub-linear in \(n\) and \(m\) or even independent from \(n\) and \(m\), we can fix the number of epochs to a small constant if we want to get some crude results, while KronRLS and SimBoost strictly require at least \(O(n^2 + m^2)\) time to get any results. 

\section{Experiments}

\subsection{Baseline methods} 

There are two baseline methods used as comparisons: SimBoost \cite{He2017} and KronRLS \cite{Pahikkala2014}, which are state-of-the-art methods for the DTI regression task.

\paragraph{SimBoost}
Simboost predicts continuous DTI values using gradient boosting regression trees. Each drug-target pair corresponds to a continuous DTI value, and the authors defined 3 types of features to characterize the drug-target pairs: type 1 features for individual entities (drugs or targets); type 2 features, derived from the drug similarity networks and target similarity networks; type 3 features, which are derived from drug-target interaction network. The 3 types of features are concatenated to form a feature vector.

Let $x_i \in R^d$ denote the vector of features for the 
\textit{i}-th drug-target pair, while $x_i \in R$ is its binding affinity. The score $\hat{y}_i$ predicted for input $x_i$ is computed as follows:

\[\hat{y}_i=\phi(x_i)=\sum_{k=1}^{K} f_k(x_i), f_k \in F\]

where $K$ is the number of regression trees and $F$ is the space of possible trees.The algorithm learns the set of trees $f_k$.

SimBoost cannot handle cold-start problems, which means it does not work for pairs in the test set with a drug or target that is absent from the training set.

\paragraph{KronRLS}
KronRLS stands for Kronecker Regularized Least Squares. It learns a prediction function $f(x)$ for drug-target pairs. It could use some similarity measure between two drug-target pairs $x$ and $x_i$, and $f(x)$ is constructed as a linear combination of the similarity values. The algorithm learns the coefficients of this linear combination from the training data.
 
Unlike SimBoost, KronRLS is applicable to cold-start problems.

\subsection{Non-proteochemometric (Compound-Only) DNN methods}
\label{subsec: non-prot}
To investigate the usefulness of including protein feature vector (PSC in this paper) in PADME, we implemented a version of PADME with only compound information as input. Different from the full PADME, this version has one output unit for each specific target protein, resulting in a network structure similar to that of \cite{ma2015}. Similar to PADME, we considered ECFP and GraphConv variants of this DNN model. We call these PADME versions Compound-Only DNNs later in this paper.

\subsection{Datasets and tools}
\label{subsec: datasets}
Similar to He et al. \cite{He2017}, we used kinase inhibitor datasets. Following its naming convention, we call them Davis dataset \cite{Davis2011}, Metz dataset \cite{metz2011} and KIBA dataset \cite{tang2014}, respectively. However, the versions of these datasets curated by Pahikkala et al. \cite{Pahikkala2014} that He et al. \cite{He2017} used was slightly different from the original dataset, and Pahikkala et al. did not give the corresponding justifications. We thus used the data provided by the respective original authors, then preprocessed them ourselves as described in the Supporting Information. We assume the observations within each dataset are under the same experimental settings. Metz dataset contained lots of imprecise values, which we discarded in the preprocessing step.

Because of the limitations of SimBoost and KronRLS, we filtered the datasets. The original KIBA dataset contains 52498 compounds, a large proportion of which only have the interaction values with very few proteins. Considering the huge compound similarity matrix required and the time-consuming matrix factorization used in SimBoost, it would be infeasible to work directly on the original KIBA dataset. Thus, we had to filter it rather aggressively so that the size becomes more manageable. We chose a threshold of 6 (drugs and targets with no more than 6 observations are removed), more lenient than the threshold of 10 used in He et al. \cite{He2017}, aiming at a reduction of the unfair advantages that SimBoost can gain by keeping only the denser submatrix of the interaction matrix.

For the Metz and Davis datasets, as SimBoost cannot handle cold drug/target problem, we had to ensure that in creating Cross-Validation folds, each drug or target appear in at least 2 folds, thus those drugs/targets with no more than 1 observation are discarded. 

We also used the ToxCast dataset \cite{EPA2015}, containing a much larger variety of proteins \cite{EPA2016}. It contains toxicology data obtained from high-throughput \emph{in vitro} screening of chemicals, mainly measured in \(AC_{50}\), which means the concentration at half of the maximum activity. The prepared dataset (see Supporting Information) contains observations for 530605 drug-target pairs. Its large size and coverage of diverse protein types allow us to test the robustness and scalability of computational models for DTI prediction. After the preprocessing, it still contains a total of 672 assays, compared to single assay/interaction strength measurement of the other 3 datasets. Some of those assays are closely related, but most of them are different from each other. Because it contains so many heterogeneous endpoints, we manually grouped those assays into 61 different measurements for interaction strength based on assay type, such that observations in each measurement are reasonably homogeneous, also increasing the number of observations for each measurement endpoint. The number of observations in each measurement range from $\mathtt{\sim}$290 to $\mathtt{\sim}$160,000. For the ToxCast dataset, we constructed multi-task networks, in which each measurement corresponds to a neuron in the output layer. As KronRLS and SimBoost are both single-task models, to evaluate the performance of those two models on the ToxCast dataset, one must train 61 models for each of them, which would be an extraordinarily tedious job, so we did not run the SimBoost and KronRLS models on ToxCast. This indicates PADME is not only more scalable in the number of drugs/targets, but also much more scalable in the number of endpoints. As ToxCast does not have the bottlenecks imposed by KronRLS and SimBoost, we did not filter it. 

Please refer to table \ref{tab1} for the sizes of the datasets after filtering.

\begin{table}[ht]
\begin{adjustwidth}{0in}{0in} 
\centering
\caption{\color{Gray}Dataset sizes after filtering.}
\begin{tabular}{|c|c|c|c|}
\hline
\thead{Dataset} & \thead{Number of drugs \\(compounds)} & \thead{Number of target \\ proteins} & \thead{Total number of \\ drug-target pairs used}\\ \hline
Davis & 72 & 442 & 31824 \\ \hline
Metz & 1423 & 170 & 35259 \\ \hline
KIBA & 3807 & 408 & 160296 \\ \hline
\makecell{ToxCast \\ (No filtering)} & 7657 & 335 & 530605 \\ \hline

\end{tabular}
\label{tab1}
\end{adjustwidth}
\end{table}

We applied the same numerical transformation as He et al. \cite{He2017} to the datasets: \(transformed= 4 - log_{10}(original)\). For the ToxCast dataset, we changed the inactive value from 1,000,000 to 1,000, so that there would be no large gaps in the distribution after transforming the data.

The model was constructed based on the implementation of the DeepChem python package \cite{ramsundar2019}, in which RDKit \cite{rdkit} was used; the networks were constructed using TensorFlow 1.3 \cite{tensorflow2016}. In the practical application, PADME takes SMILES representations of the candidate drug as part of the input, which are transformed into graph representations or ECFP by the program. PSC was obtained independently from this process: we used the propy python package \cite{cao2013} to generate PSC descriptors, and manually added a binary entry indicating phosphorylation. Afterwards, PSC was saved in a standalone file, which the program reads into the memory in the runtime.

The experiments were conducted on a Linux server with 8 Nvidia Geforce GTX 1080Ti graphics cards, among which 4 were used. The server has 40 logical CPU cores and 256 GB of RAM. A computer with less than 110 GB RAM might not be able to perform cross-validation for the ToxCast dataset using GraphConv-based PADME.

\subsection{Experimental Design}
\label{sec:design}

To examine PADME's prediction power, we used cross-validation (CV), which is the convention of the prior researches, also because we believe the comprehensive coverage of the whole dataset will offer a more thorough evaluation of the model's performance, rather than only using 1 hold-out test set. To measure the performance of the model under different settings, multiple CV splitting schemes were employed to evaluate the predictions of the models trained from the training sets against the known interaction strengths in the test sets.  The performances of PADME-ECFP and PADME-GraphConv were compared against each other under identical settings.

We performed 5-fold CV. For SimBoost to work, every compound (candidate drug) or target must be present in at least 2 folds, this splitting scheme is called ``warm split'' in this paper. There are no such restrictions for KronRLS, since it can handle cold-start data. Since we did not run SimBoost on ToxCast data, there is no need to perform warm-split on it, we then used random split in that case. If we force a warm split on the ToxCast dataset, a filter threshold of 1 must be used to reduce the size of the dataset, which is undesirable. As cold-start prediction is an important objective in DTI prediction (and an advantage of PADME), we also included cold-splitting in constructing the cross-validation folds, such that all compounds (candidate drugs) in the test fold are absent from the training fold (cold-drug split), or all targets in the test fold are absent from the training fold (cold-target split). Also, similar to \cite{mayr2016}, we also implemented a cold-drug cluster split, using single-linkage clustering with Tanimoto similarity to create compound clusters. Compounds whose ECFP4 fingerprint had higher similarity than 0.7 were assigned to the same cluster. Compounds belonging to the same cluster were assigned to same folds, so that compounds in the validation fold would not be similar to those in the training fold. The cold-drug cluster split can prevent the performance estimation from being overly optimistic.  Though \cite{Pahikkala2014} suggested another splitting scheme which results in simultaneous cold-drug and cold-target in each validation fold, as it greatly decreases the size of the training set in each fold (4/9 of the original data instead of 4/5 in other splitting schemes), we decided that it would cause unfair comparison and did not use it.

For every dataset, we performed four types of CV splitting (warm, cold-target, cold-drug, cold-drug cluster), and for every CV splitting scheme, we evaluated the prediction performances of the applicable models (KronRLS and PADME for all splitting schemes, SimBoost for warm splits only). To reduce the random effects, we repeated the splitting several times for each splitting scheme on all the datasets and calculated the average values of the evaluation metrics of the prediction results across the splits. The Compound-Only DNNs (as mentioned in \autoref{subsec: non-prot}) take only compound information as input and predict the response for multiple proteins simultaneously. Therefore, they cannot handle cold-target scenarios, and it is unnatural to test them in a warm-split scenario. We only use them to compare against PADME in cold-drug and cold-drug cluster splits.

Not only do we have multiple splitting methods, we also used multiple model settings and evaluation metrics. For each of PADME-ECFP and PADME-GraphConv, a single-task network was trained for every splitting scheme of every dataset, except ToxCast, for which we constructed a multi-task network with 61 output neurons to avoid the complexity caused by 61 separate single-task networks. 

We also wanted to investigate whether PADME can predict the ordering of the interaction strengths correctly, so in addition to metrics focusing on value correctness (RMSE (Root Mean Squared Error) and \(R^2\)), we also used metrics focusing on order correctness, like concordance index (CI). Using CI as a metric in cheminformatics setting was proposed by Pahikkala et al. \cite{Pahikkala2014}. It measures the probability of correctly ordering the non-equal pairs in the dataset, ranging across [0, 1], with bigger values indicating better results. If you use the same value (e.g. mean value of the training set) as the predicted results across the test set, the CI would be 0.5. We note that the CI neglects the magnitude of values while focusing on the pairwise comparison, and it does not consider the prediction correctness for data points that truly have values equal to each other. Thus, CI should be used alongside other metrics like RMSE. However, in virtual screening, we are typically only interested in the top predictions, so that the drawback of neglecting the magnitude is not a big concern.

\begin{wrapfigure}{l}{80mm}
\includegraphics[width=80mm]{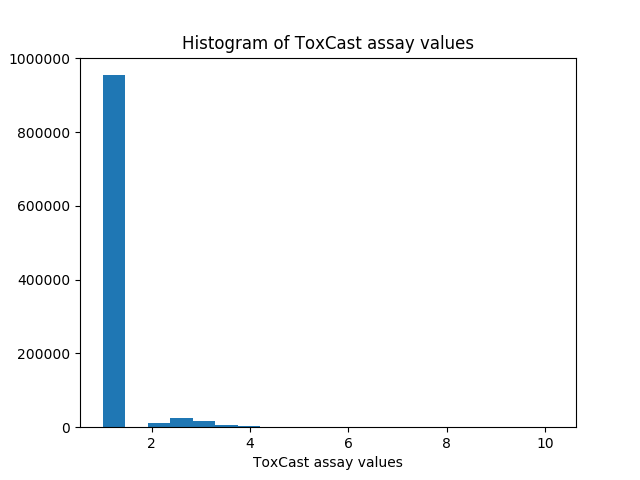}
\captionsetup{labelformat=empty, width=0.2\textwidth} 
\caption{} 
\label{toxcast_hist}
\end{wrapfigure}

\marginpar{
\vspace{.7cm} 
\color{Gray} 
\textbf{Figure \ref{toxcast_hist}.} 
The histogram of the distribution of the negative log transformed ToxCast measurement results. The majority (over 94\%) are concentrated at one inactive value.
}

To improve the readability of the reported results for the ToxCast dataset, the performance metrics are averaged across the 61 different measurements, weighted by the number of records for each of the measurements, so the results reported for the ToxCast dataset look the same as other datasets with single endpoints. 

As an exploratory analysis of the datasets, we found the ToxCast to be special. As shown in Figure \ref{toxcast_hist}, the transformed ToxCast dataset is extremely concentrated at a value of 1 which corresponds to no interaction. This led us to ignore the \(R^2\) values for this dataset: because \(R^2\) is sensitive to the overall departure of the predicted values from the true values, we argue that the huge concentration of values has rendered \(R^2\) uninformative in measuring the performance of the model on the ToxCast dataset. This concentration of values also makes RMSE less informative than it otherwise would be (since one can blindly guess inactive values for all and still get pretty good RMSE), so we argue that CI is the most useful metric in the ToxCast dataset prediction evaluation. This pronounced imbalance in the dataset caused us to consider balancing it through oversampling (see supporting information).

Following the principle of parsimony, we wanted to use a minimal number of hyperparameter sets wherever possible, to keep the time and computational costs manageable. If, instead, we do one hyperparameter tuning to get the hyperparameters for each CV iteration, to ensure a reasonable coverage of parameter space, it would have taken well over a month to run a CV for a dataset due to the intrinsic complexity of DNN models involving protein information, which would have been unrealistic, both for us and future users. So we cannot use the nested/double cross-validation as used in \cite{baumann2014} and \cite{mayr2018}. Also, since the datasets are not very large for deep learning, we wanted to maximize the training set size in each iteration. Thus, in our hyperparameter tuning process, we randomly selected 90\% of the dataset to be the training set, the remaining 10\% to be the validation set, and the validation set was used for determining the best set of hyperparameters. Then in the 5-fold CV splits, we excluded the aforementioned validation set elements from each validation fold, but the validation set elements are retained in the training folds, thus in each CV iteration, the training folds are 80\% the size of the dataset, while the validation fold is 18\% the size of the dataset. This simultaneously makes the training folds as large as possible, and avoids bias in evaluation.

To efficiently tune the hyperparameters (like dropout rates, batch size, learning rate, number of layers, nodes per layer, etc.) for both PADME models and Compound-Only DNN models, we used Bayesian Optimization \cite{shahriari2016} implemented by the Python package pygpgo \cite{pygpgo}. We also used early stopping to determine the number of training epochs needed. To guide early stopping, we used \(mean(RMSE)-mean(CI)\) calculated on the validation set as the composite score to be minimized. The training and validation sets used for hyperparameter searching are split randomly from the original dataset without any concern for warm or cold drug/target splits. We only store one optimal set of hyperparameters per \emph{(dataset, PADME variant)} pair, which were then used for all CV settings for that \emph{(dataset, PADME variant)} pair. Note that, for simplicity and to examine the robustness of PADME, the set of hyperparameters found in the random splitting was used in all CV settings, including those with cold-drug split and cold-target split, though we believe better CV results could be achieved if the hyperparameter searching processes are specifically designed for that CV fold split scheme, e.g., for cold-target CV folds, we could use the hyperparameters found by running the Bayesian Optimization on cold-target splitted datasets.

The resulting networks typically have 2 or 3 fully-connected layers connecting the CIV to the output unit, with thousands of neurons in each of the layers. Each fully-connected layer is batch-normalized. 

\subsection{Experimental Results}
\subsubsection{Quantitative Results}
Based on the experimental design in \autoref{sec:design}, we obtained the quantitative results for PADME. In \cref{tab_rmse,tab_ci,tab_r2}, the bold numbers indicate the best values attained for each setting. The sample standard deviations of CV mean results are also calculated, based on which we performed two-sample t-tests with unequal variances. For each t-test, the null hypothesis was that the mean of the CV results of the model is not worse than the best PADME result (boldfaced ones), while the alternative was that the model was worse than the best PADME model. For RMSE, worse means larger, while for CI or \(R^2\), worse means smaller. The p-values are reported. We observe that the two versions of PADME dominate the other methods\footnote{Note that the SimBoost results reported here are considerably worse than the results reported in their original paper. It is because we have examined their source code and found they calculated MSE but reported it as RMSE.}, including the Compound-Only DNN models though to a lesser degree, across all datasets and splits for all evaluation metrics. 

We note the following exceptions. SimBoost outperforms PADME-GraphConv on the Metz dataset, which could be due to the small dataset size: PADME-GraphConv could be overfitting for Metz data, while SimBoost uses gradient boosting trees, a machine learning model better suited for small datasets than Deep Neural Networks. Because it does not use MGC, PADME-ECFP has a much smaller network than PADME-GraphConv, which may explain why the former performs slightly better on the Metz dataset. However, we do not observe the same phenomenon on the Davis dataset, which has a similar size and even fewer entities. Comparing PADME-ECFP against Compound-Only ECFP and PADME-GraphConv against Compound-Only GraphConv, we observe that PADME consistently performs better in Concordance Index, for example, they outperform Compound-Only DNNs by around 10\% or even more in Concordance Index on the ToxCast dataset. But in RMSE and \(R^2\), Compound-Only DNNs sometimes perform similarly to the PADME models in Davis and KIBA datasets, or even \textbf{insignificantly} outperform PADME models. In general, it seems that PADME models are better than others, but more so in predicting the order, as reflected in CI. Besides, Compound-Only DNNs outperform KronRLS in general, except the Compound-Only GraphConv on the Davis dataset.

\begin{table}[!hp]
\centering
\caption{The regression performance across the datasets measured in RMSE (smaller is better), averaged across independent repetitions of CV. The mean RMSE are enclosed in square brackets; sample standard deviations are also reported. The best results in the PADME models are boldfaced, and one-sided two-sample t-tests are conducted against them. The \textcolor{blue}{blue} values are insignificantly bigger (worse) (p $>$ 0.05) than the boldfaced values, while the \textcolor{orange}{orange} ones are insignificantly smaller (better) (p $<$ 0.95) than them. The uncolored ones are significantly worse than the boldfaced values.}
\resizebox{\textwidth}{!}{
\begin{tabular}{|c|c|c|c|c|c|c|c|c|}
\hline

{} & {} & {} & \multicolumn{6}{c|}{RMSE}\\ \hline
Dataset & \makecell{Cross Validation \\ Splitting type} & \makecell{Value \\ Type} & \makecell{PADME-\\ECFP} & \makecell{PADME-\\GraphConv} & \makecell{SimBoost} & \makecell{KronRLS} & \makecell{Compound-Only\\ ECFP} & \makecell{Compound-Only\\ GraphConv} \\ \hline
\multirow{12}{*}{Davis} & \multirow{3}{*}{Warm} & mean & \textbf{[0.4287]} & \textcolor{blue}{[0.4313]} & [0.4820] & [0.5729] & \NA & \NA \\\cline{3-9}
& & std & 0.0029 & 0.0029 & 0.0019 & 0.0051 & \NA & \NA \\\cline{3-9}
& & p-val & \NA & 0.0969 & 4.25E-9 & 2.97E-12 & \NA & \NA \\\cline{2-9}
& \multirow{3}{*}{Cold Drug} & mean & \textbf{[0.8054]} & [0.8280] & \NA & [0.8405] & \textcolor{orange}{[0.8038]} & [0.8505] \\\cline{3-9}
& & std & 0.0118 & 0.0192 & \NA & 0.0210 & 0.0148 & 0.0238 \\\cline{3-9}
& & p-val & \NA & 0.0312 & \NA & 0.0041 & 0.5733 & 0.0047 \\\cline{2-9}
& \multirow{3}{*}{\makecell{Cold Drug \\ Cluster}} & mean & \textbf{[0.7671]} & [0.7922] & \NA & [0.8368] & \textcolor{blue}{[0.8040]} & [0.8351] \\\cline{3-9}
& & std & 0.0171 & 0.0228 & \NA & 0.0382 & 0.0435 & 0.0144 \\\cline{3-9}
& & p-val & \NA & 0.0435 & \NA & 0.0024 & 0.0675 & 7.86E-5 \\\cline{2-9}
& \multirow{3}{*}{Cold Target} & mean & \textbf{[0.5639]} & [0.5748] & \NA & [0.6596] & \NA & \NA \\\cline{3-9}
& & std & 0.0065 & 0.0080 & \NA & 0.0020 & \NA & \NA \\\cline{3-9}
& & p-val & \NA & 0.0229 & \NA & 6.5E-7 & \NA & \NA \\\hline
\multirow{12}{*}{Metz} & \multirow{3}{*}{Warm} & mean & \textbf{[0.5556]} & [0.6100] & [0.5813] & [0.7813] & \NA & \NA \\\cline{3-9}
& & std & 0.0022 & 0.0111 & 0.0016 & 3.87E-4 & \NA & \NA \\\cline{3-9}
& & p-val & \NA & 1.4E-4 & 4.29E-8 & 4.42E-10 & \NA & \NA \\\cline{2-9}
& \multirow{3}{*}{Cold Drug} & mean & \textbf{[0.7119]} & [0.7533] & \NA & [0.7843] & [0.7738] & [0.7775] \\\cline{3-9}
& & std & 0.0016 & 0.0086 & \NA & 0.0052 & 0.0146 & 0.0045 \\\cline{3-9}
& & p-val & \NA & 1.6E-4 & \NA & 2.41E-8 & 3.10E-4 & 3.16E-7 \\\cline{2-9}
& \multirow{3}{*}{\makecell{Cold Drug \\ Cluster}} & mean & \textbf{[0.7770]} & [0.8099] & \NA & [0.8315] & [0.8250] & [0.8386] \\\cline{3-9}
& & std & 0.0115 & 0.0089 & \NA & 0.0054 & 0.0075 & 0.0051 \\\cline{3-9}
& & p-val & \NA & 5.8E-4 & \NA & 5.81E-5 & 5.56E-5 & 2.94E-5 \\\cline{2-9}
& \multirow{3}{*}{Cold Target} & mean & \textbf{[0.7905]} & [0.8239] & \NA & [0.8989] & \NA & \NA \\\cline{3-9}
& & std & 0.0127 & 0.0107 & \NA & 0.0101 & \NA & \NA \\\cline{3-9}
& & p-val & \NA & 0.0011 & \NA & 2.56E-7 & \NA & \NA \\\hline
\multirow{12}{*}{KIBA} & \multirow{3}{*}{Warm} & mean & [0.4334] & \textbf{[0.4247]} & [0.4689] & [0.6566] & \NA & \NA \\\cline{3-9}
& & std & 0.0069 & 0.0027 & 0.0010 & 1.74E-4 & \NA & \NA \\\cline{3-9}
& & p-val & 0.0405 & \NA & 4.93E-6 & 2.01E-7 & \NA & \NA \\\cline{2-9}
& \multirow{3}{*}{Cold Drug} & mean & \textbf{[0.6007]} & [0.6444] & \NA & [0.7024] & [0.6319] & [0.6421] \\\cline{3-9}
& & std & 0.0036 & 0.0149 & \NA & 0.0024 & 0.0045 & 0.0024 \\\cline{3-9}
& & p-val & \NA & 0.0040 & \NA & 1.85E-8 & 3.93E-6 & 2.42E-6 \\\cline{2-9}
& \multirow{3}{*}{\makecell{Cold Drug \\ Cluster}} & mean & \textbf{[0.7132]} & \textcolor{blue}{[0.7263]} & \NA & [0.7536] & \textcolor{orange}{[0.7029]} & \textcolor{blue}{[0.7196]} \\\cline{3-9}
& & std & 0.0270 & 0.0224 & \NA & 0.0039 & 0.0064 & 4.71E-4 \\\cline{3-9}
& & p-val & \NA & 0.2409 & \NA & 0.0285 & 0.7459 & 0.3329 \\\cline{2-9}
& \multirow{3}{*}{Cold Target} & mean & \textcolor{blue}{[0.6226]} & \textbf{[0.6225]} & \NA & [0.6811] & \NA & \NA \\\cline{3-9}
& & std & 0.0035 & 0.0058 & \NA & 0.0082 & \NA & \NA \\\cline{3-9}
& & p-val & 0.4867 & \NA & \NA & 2.37E-5 & \NA & \NA \\\hline
\multirow{12}{*}{ToxCast} & \multirow{3}{*}{Warm} & mean & \textbf{[0.4049]} & [0.4092] & \NA & \NA & \NA & \NA \\\cline{3-9}
& & std & 0.0011 & 0.0013 & \NA & \NA & \NA & \NA \\\cline{3-9}
& & p-val & \NA & 0.0012 & \NA & \NA & \NA & \NA \\\cline{2-9}
& \multirow{3}{*}{Cold Drug} & mean & \textbf{[0.4447]} & \textcolor{blue}{[0.4448]} & \NA & \NA & [0.4550] & [0.4682] \\\cline{3-9}
& & std & 6.37E-4 & 3.3E-4 & \NA & \NA & 0.0012 & 0.0036 \\\cline{3-9}
& & p-val & \NA & 0.4343 & \NA & \NA & 1.03E-6 & 4.42E-5 \\\cline{2-9}
& \multirow{3}{*}{\makecell{Cold Drug \\ Cluster}} & mean & [0.4480] & \textbf{[0.4476]} & \NA & \NA & [0.4509] & [0.4566] \\\cline{3-9}
& & std & 0.0006 & 0.0014 & \NA & \NA & 9.53E-4 & 0.0025 \\\cline{3-9}
& & p-val & 0.3264 & \NA & \NA & \NA & 0.0048 & 1.82E-4 \\\cline{2-9}
& \multirow{3}{*}{Cold Target} & mean & \textbf{[0.4794]} & \textcolor{blue}{[0.4896]} & \NA & \NA & \NA & \NA \\\cline{3-9}
& & std & 0.0089 & 0.0120 & \NA & \NA & \NA & \NA \\\cline{3-9}
& & p-val & \NA & 0.1133 & \NA & \NA & \NA & \NA \\\hline

\end{tabular}}
\label{tab_rmse}
\end{table}

\begin{table}[!hp]
\centering
\caption{The regression performance across the datasets measured in Concordance Index (larger is better), averaged across independent repetitions of CV. Similar to \cref{tab_rmse}, the mean CI are enclosed in square brackets; sample standard deviations are also reported. One-sided two-sample t-tests are conducted against the best PADME models. The \textcolor{blue}{blue} values are insignificantly smaller (worse) (p $>$ 0.05) than the boldfaced values, while the \textcolor{orange}{orange} ones are insignificantly larger (better) (p $<$ 0.95). The uncolored ones are significantly worse than the boldfaced values.}
\resizebox{\textwidth}{!}{
\begin{tabular}{|c|c|c|c|c|c|c|c|c|}
\hline

{} & {} & {} & \multicolumn{6}{c|}{Concordance Index}\\ \hline
Dataset & \makecell{Cross Validation \\ Splitting type} & \makecell{Value \\ Type} & \makecell{PADME-\\ECFP} & \makecell{PADME-\\GraphConv} & \makecell{SimBoost} & \makecell{KronRLS} & \makecell{Compound-Only\\ ECFP} & \makecell{Compound-Only\\ GraphConv} \\ \hline
\multirow{12}{*}{Davis} & \multirow{3}{*}{Warm} & mean & \textcolor{blue}{[0.9034]} & \textbf{[0.9040]} & [0.8871] & [0.8758] & \NA & \NA \\\cline{3-9}
& & std & 0.0020 & 0.0012 & 5.97E-4 & 0.0015 & \NA & \NA \\\cline{3-9}
& & p-val & 0.2780 & \NA & 1.35E-7 & 2.97E-11 & \NA & \NA \\\cline{2-9}
& \multirow{3}{*}{Cold Drug} & mean & \textbf{[0.7120]} & \textcolor{blue}{[0.7099]} & \NA & [0.6924] & \textcolor{blue}{[0.7027]} & [0.6668] \\\cline{3-9}
& & std & 0.0026 & 0.0139 & \NA & 0.0117 & 0.0162 & 0.0187 \\\cline{3-9}
& & p-val & \NA & 0.3801 & \NA & 0.0042 & 0.1361 & 0.0026 \\\cline{2-9}
& \multirow{3}{*}{\makecell{Cold Drug \\ Cluster}} & mean & \textbf{[0.7238]} & \textcolor{blue}{[0.7190]} & \NA & [0.6800] & \textcolor{blue}{[0.6994]} & [0.6828] \\\cline{3-9}
& & std & 0.0094 & 0.0096 & \NA & 0.0421 & 0.0265 & 0.0215 \\\cline{3-9}
& & p-val & \NA & 0.2208 & \NA & 0.0254 & 0.0547 & 0.0047 \\\cline{2-9}
& \multirow{3}{*}{Cold Target} & mean & \textbf{[0.8538]} & [0.8428] & \NA & [0.8075] & \NA & \NA \\\cline{3-9}
& & std & 0.0034 & 0.0031 & \NA & 0.0027 & \NA & \NA \\\cline{3-9}
& & p-val & \NA & 3.70E-4 & \NA & 8.45E-9 & \NA & \NA \\\hline
\multirow{12}{*}{Metz} & \multirow{3}{*}{Warm} & mean & \textbf{[0.8065]} & [0.7931] & [0.7944] & [0.7485] & \NA & \NA \\\cline{3-9}
& & std & 0.0012 & 0.0016 & 6.92E-4 & 6.98E-4 & \NA & \NA \\\cline{3-9}
& & p-val & \NA & 3.49E-7 & 5.05E-7 & 3.71E-11 & \NA & \NA \\\cline{2-9}
& \multirow{3}{*}{Cold Drug} & mean & \textbf{[0.7432]} & [0.7384] & \NA & [0.7092] & [0.7110] & [0.7197] \\\cline{3-9}
& & std & 0.0021 & 0.0013 & \NA & 0.0021 & 0.0033 & 0.0037 \\\cline{3-9}
& & p-val & \NA & 0.0018 & \NA & 5.63E-10 & 2.60E-7 & 6.11E-6 \\\cline{2-9}
& \multirow{3}{*}{\makecell{Cold Drug \\ Cluster}} & mean & \textbf{[0.7158]} & \textcolor{blue}{[0.7132]} & \NA & [0.6818] & [0.6878] & [0.6966] \\\cline{3-9}
& & std & 0.0048 & 0.0014 & \NA & 0.0037 & 0.0018 & 0.0026 \\\cline{3-9}
& & p-val & \NA & 0.1485 & \NA & 1.02E-6 & 2.82E-5 & 9.20E-5 \\\cline{2-9}
& \multirow{3}{*}{Cold Target} & mean & [0.6961] & \textbf{[0.7099]} & \NA & [0.6470] & \NA & \NA \\\cline{3-9}
& & std & 0.0076 & 0.0031 & \NA & 0.0048 & \NA & \NA \\\cline{3-9}
& & p-val & 0.0058 & \NA & \NA & 7.82E-10 & \NA & \NA \\\hline
\multirow{12}{*}{KIBA} & \multirow{3}{*}{Warm} & mean & [0.8577] & \textbf{[0.8616]} & [0.8405] & [0.7831] & \NA & \NA \\\cline{3-9}
& & std & 0.0011 & 0.0014 & 1.35E-4 & 3.26E-4 & \NA & \NA \\\cline{3-9}
& & p-val & 0.0028 & \NA & 3.60E-5 & 2.59E-7 & \NA & \NA \\\cline{2-9}
& \multirow{3}{*}{Cold Drug} & mean & \textbf{[0.7742]} & [0.7524] & \NA & [0.6890] & [0.7405] & [0.7356] \\\cline{3-9}
& & std & 0.0011 & 0.0032 & \NA & 0.0014 & 0.0027 & 0.0023 \\\cline{3-9}
& & p-val & \NA & 1.64E-4 & \NA & 9.49E-11 & 3.05E-7 & 2.66E-8 \\\cline{2-9}
& \multirow{3}{*}{\makecell{Cold Drug \\ Cluster}} & mean & \textbf{[0.7465]} & [0.7190] & \NA & [0.6654] & [0.7074] & [0.7068] \\\cline{3-9}
& & std & 0.0019 & 0.0030 & \NA & 0.0038 & 0.0034 & 0.0023 \\\cline{3-9}
& & p-val & \NA & 9.77E-6 & \NA & 5.43E-9 & 1.37E-7 & 9.90E-9 \\\cline{2-9}
& \multirow{3}{*}{Cold Target} & mean & \textcolor{blue}{[0.7684]} & \textbf{[0.7687]} & \NA & [0.7122] & \NA & \NA \\\cline{3-9}
& & std & 0.0020 & 0.0029 & \NA & 0.0045 & \NA & \NA \\\cline{3-9}
& & p-val & 0.4210 & \NA & \NA & 1.82E-6 & \NA & \NA \\\hline
\multirow{12}{*}{ToxCast} & \multirow{3}{*}{Warm} & mean & [0.7908] & \textbf{[0.7963]} & \NA & \NA & \NA & \NA \\\cline{3-9}
& & std & 0.0041 & 8.62E-4 & \NA & \NA & \NA & \NA \\\cline{3-9}
& & p-val & 0.0380 & \NA & \NA & \NA & \NA & \NA \\\cline{2-9}
& \multirow{3}{*}{Cold Drug} & mean & [0.7196] & \textbf{[0.7329]} & \NA & \NA & [0.6692] & [0.6414] \\\cline{3-9}
& & std & 0.0062 & 0.0027 & \NA & \NA & 0.0041 & 0.0068 \\\cline{3-9}
& & p-val & 0.0082 & \NA & \NA & \NA & 1.25E-8 & 2.18E-7 \\\cline{2-9}
& \multirow{3}{*}{\makecell{Cold Drug \\ Cluster}} & mean & [0.7161] & \textbf{[0.7290]} & \NA & \NA & [0.6440] & [0.6192] \\\cline{3-9}
& & std & 0.0012 & 0.0032 & \NA & \NA & 0.0045 & 0.0061 \\\cline{3-9}
& & p-val & 0.0010 & \NA & \NA & \NA & 3.27E-9 & 1.20E-8 \\\cline{2-9}
& \multirow{3}{*}{Cold Target} & mean & [0.6752] & \textbf{[0.6979]} & \NA & \NA & \NA & \NA \\\cline{3-9}
& & std & 0.0108 & 0.0116 & \NA & \NA & \NA & \NA \\\cline{3-9}
& & p-val & 0.0144 & \NA & \NA & \NA & \NA & \NA \\\hline

\end{tabular}}
\label{tab_ci}
\end{table}

\begin{table}[!hp]
\centering
\caption{The regression performance across the datasets measured in \(R^2\) (larger is better), averaged across independent repetitions of CV. Similar to \cref{tab_rmse}, one-sided two-sample t-tests are conducted against the best PADME models. The \textcolor{blue}{blue} values are insignificantly smaller (worse) (p $>$ 0.05) than the boldfaced values, while the \textcolor{orange}{orange} ones are insignificantly larger (better) (p $<$ 0.95). The uncolored ones are significantly worse than the boldfaced values.}
\resizebox{\textwidth}{!}{
\begin{tabular}{|c|c|c|c|c|c|c|c|c|}
\hline
{} & {} & {} & \multicolumn{6}{c|}{\(R^2\)}\\ \hline
Dataset & \makecell{Cross Validation \\ Splitting type} & \makecell{Value \\ Type} & \makecell{PADME-\\ECFP} & \makecell{PADME-\\GraphConv} & \makecell{SimBoost} & \makecell{KronRLS} & \makecell{Compound-Only\\ ECFP} & \makecell{Compound-Only\\ GraphConv} \\ \hline
\multirow{12}{*}{Davis} & \multirow{3}{*}{Warm} & mean & \textbf{[0.7652]} & [0.7614] & [0.7031] & [0.5801] & \NA & \NA \\\cline{3-9}
& & std & 0.0029 & 0.0031 & 0.0023 & 0.0077 & \NA & \NA \\\cline{3-9}
& & p-val & \NA & 0.0402 & 2.79E-10 & 2.42E-10 & \NA & \NA \\\cline{2-9}
& \multirow{3}{*}{Cold Drug} & mean & \textbf{[0.1439]} & [0.0851] & \NA & [0.0478] & \textcolor{blue}{[0.1358]} & [0.0612] \\\cline{3-9}
& & std & 0.0335 & 0.0499 & \NA & 0.0592 & 0.0446 & 0.0606 \\\cline{3-9}
& & p-val & \NA & 0.0325 & \NA & 0.0047 & 0.3762 & 0.0178 \\\cline{2-9}
& \multirow{3}{*}{\makecell{Cold Drug \\ Cluster}} & mean & \textbf{[0.2253]} & [0.1631] & \NA & [0.0540] & \textcolor{blue}{[0.1174]} & [0.0991] \\\cline{3-9}
& & std & 0.0340 & 0.0520 & \NA & 0.0931 & 0.1172 & 0.0321 \\\cline{3-9}
& & p-val & \NA & 0.0304 & \NA & 0.0024 & 0.0545 & 1.57E-4 \\\cline{2-9}
& \multirow{3}{*}{Cold Target} & mean & \textbf{[0.5915]} & [0.5741] & \NA & [0.4393] & \NA & \NA \\\cline{3-9}
& & std & 0.0087 & 0.0127 & \NA & 0.0064 & \NA & \NA \\\cline{3-9}
& & p-val & \NA & 0.0195 & \NA & 2.02E-9 & \NA & \NA \\\hline
\multirow{12}{*}{Metz} & \multirow{3}{*}{Warm} & mean & \textbf{[0.6654]} & [0.5961] & [0.6323] & [0.3355] & \NA & \NA \\\cline{3-9}
& & std & 0.0027 & 0.0153 & 0.0020 & 6.15E-4 & \NA & \NA \\\cline{3-9}
& & p-val & \NA & 2.1E-4 & 2.56E-8 & 1.18E-10 & \NA & \NA \\\cline{2-9}
& \multirow{3}{*}{Cold Drug} & mean & \textbf{[0.4477]} & [0.3813] & \NA & [0.3285] & [0.3471] & [0.3430] \\\cline{3-9}
& & std & 0.0024 & 0.0145 & \NA & 0.0084 & 0.0241 & 0.0114 \\\cline{3-9}
& & p-val & \NA & 2E-4 & \NA & 2.59E-8 & 3.41E-4 & 8.93E-6 \\\cline{2-9}
& \multirow{3}{*}{\makecell{Cold Drug \\ Cluster}} & mean & \textbf{[0.3390]} & [0.2779] & \NA & [0.2416] & [0.2457] & [0.2241] \\\cline{3-9}
& & std & 0.0183 & 0.0187 & \NA & 0.0080 & 0.0158 & 0.0099 \\\cline{3-9}
& & p-val & \NA & 4E-4 & \NA & 3.93E-5 & 1.46E-5 & 7.22E-6 \\\cline{2-9}
& \multirow{3}{*}{Cold Target} & mean & \textbf{[0.3185]} & [0.2562] & \NA & [0.1130] & \NA & \NA \\\cline{3-9}
& & std & 0.0214 & 0.0207 & \NA & 0.0195 & \NA & \NA \\\cline{3-9}
& & p-val & \NA & 8E-4 & \NA & 6.30E-8 & \NA & \NA \\\hline

\multirow{12}{*}{KIBA} & \multirow{3}{*}{Warm} & mean & [0.7449] & \textbf{[0.7560]} & [0.7007] & [0.4128] & \NA & \NA \\\cline{3-9}
& & std & 0.0084 & 0.0032 & 0.0013 & 3.34E-4 & \NA & \NA \\\cline{3-9}
& & p-val & 0.0356 & \NA & 2.92E-6 & 8.54E-8 & \NA & \NA \\\cline{2-9}
& \multirow{3}{*}{Cold Drug} & mean & \textbf{[0.5093]} & [0.4352] & \NA & [0.3266] & [0.4599] & [0.4312] \\\cline{3-9}
& & std & 0.0057 & 0.0270 & \NA & 0.0048 & 0.0039 & 0.0054 \\\cline{3-9}
& & p-val & \NA & 0.0051 & \NA & 3.57E-9 & 9.92E-6 & 1.93E-7 \\\cline{2-9}
& \multirow{3}{*}{\makecell{Cold Drug \\ Cluster}} & mean & \textbf{[0.2948]} & \textcolor{blue}{[0.2793]} & \NA & \textcolor{blue}{[0.2215]} & \textcolor{orange}{[0.3267]} & \textcolor{blue}{[0.2887]} \\\cline{3-9}
& & std & 0.0687 & 0.0463 & \NA & 0.0070 & 0.0140 & 0.0042 \\\cline{3-9}
& & p-val & \NA & 0.3620 & \NA & 0.0611 & 0.7878 & 0.4354 \\\cline{2-9}
& \multirow{3}{*}{Cold Target} & mean & \textcolor{blue}{[0.4715]} & \textbf{[0.4734]} & \NA & [0.3631] & \NA & \NA \\\cline{3-9}
& & std & 0.0051 & 0.0100 & \NA & 0.0136 & \NA & \NA \\\cline{3-9}
& & p-val & 0.3752 & \NA & \NA & 1.14E-5 & \NA & \NA \\\hline

\multirow{12}{*}{ToxCast} & \multirow{3}{*}{Warm} & mean & \NA & \NA & \NA & \NA & \NA & \NA \\\cline{3-9}
& & std & \NA & \NA & \NA & \NA & \NA & \NA \\\cline{3-9}
& & p-val & \NA & \NA & \NA & \NA & \NA & \NA \\\cline{2-9}
& \multirow{3}{*}{Cold Drug} & mean & \NA & \NA & \NA & \NA & \NA & \NA \\\cline{3-9}
& & std & \NA & \NA & \NA & \NA & \NA & \NA \\\cline{3-9}
& & p-val & \NA & \NA & \NA & \NA & \NA & \NA \\\cline{2-9}
& \multirow{3}{*}{\makecell{Cold Drug \\ Cluster}} & mean & \NA & \NA & \NA & \NA & \NA & \NA \\\cline{3-9}
& & std & \NA & \NA & \NA & \NA & \NA & \NA \\\cline{3-9}
& & p-val & \NA & \NA & \NA & \NA & \NA & \NA \\\cline{2-9}
& \multirow{3}{*}{Cold Target} & mean & \NA & \NA & \NA & \NA & \NA & \NA \\\cline{3-9}
& & std & \NA & \NA & \NA & \NA & \NA & \NA \\\cline{3-9}
& & p-val & \NA & \NA & \NA & \NA & \NA & \NA \\\hline

\end{tabular}}
\label{tab_r2}
\textsuperscript{\emph{a}} We did not report \(R^2\) for ToxCast because of its imbalanced/skewed nature.
\end{table}

It is somewhat surprising that PADME-ECFP is not outperformed by PADME-GraphConv; instead, it slightly outperforms PADME-GraphConv in many cases, though in general their performances are very close to each other. The Compound-Only ECFP models usually outperforms Compound-Only GraphConv. PADME-ECFP only takes about 23\% of the time and 45\% the space (RAM) of PADME-GraphConv in the training process and yields similar (and sometimes better) results, so PADME-ECFP is a more reasonable choice. Nonetheless, we cannot be certain that PADME-GraphConv and PADME-ECFP truly have similar performances, as there might be a better set of hyperparameters for each model that would differentiate their performances significantly. We think that the higher complexity of PADME-GraphConv introduced by the MGC network makes it harder to find a good set of hyperparameters, while it is relatively easier to find a good set of hyperparameters for PADME-ECFP which has a simpler network. This could be a possible reason why MGC cannot beat ECFP in our experiments. Thus, future researchers should continue investigating MGC and find better sets of hyperparameters in PADME-GraphConv, and perhaps propose better MGC models.

From \cref{tab_rmse,tab_ci,tab_r2} we can observe an interesting phenomenon: when there are many compounds and few targets in the training set, the cold-drug predictions tend to outperform the cold-target predictions; on the other hand, when there are many targets and few compounds, the cold-target predictions tend to be better than the cold-drug ones. We hypothesize that it is because the models can be more robust in entities (drugs or targets) with more information in the training set, thus performing better in the corresponding scenario. This trend is not only present in the PADME models, but in KronRLS as well. It seems that the models also require much more types of compounds than proteins for learning their chemical features, as can be seen from the KIBA dataset, whose cold-drug and cold-target performances are very similar, though it has 3807 compounds and only 408 proteins. And, as expected, there is a universal trend that the performance of warm splits is always better than that of cold-drug, cold-drug cluster, or cold-target splits.

The use of cold-drug clusters prevents us from overestimating the performance of the models: in the Metz and KIBA datasets, the performances for cold-drug cluster CV are noticeably worse than those for cold-drug CVs, while the performances on the Davis and ToxCast datasets stay almost the same. This could be due to the different distributions of compounds in different datasets. We suggest that future researches also employ cold-drug clusters splits in their experiments, so that a more stringent evaluation could be performed.

The fact that PADME outperforms both SimBoost and KronRLS demonstrates the power of DNN to learn complicated nonlinear relationships between drug-target pairs and interaction strength.
We were also able to show the superiority of PADME over the Compound-Only DNN models, both in the applicability of cold-target scenario and overall performance, which might suggest the improvement introduced by protein-specific descriptors (PSC in this paper). Furthermore, the results presented might be an understatement of the real performance of PADME in cold-drug and cold-target scenarios, as the training and validation sets for hyperparameter searching are randomly split, resulting in a set of hyperparameters that suit well for randomly split CV folds, but perhaps not for cold-drug and cold-target folds. This deliberately unfair comparison shows the robustness of the PADME models.

\paragraph{Applicability Domain}
Applicability Domain (AD) is an important issue in considering the usage of QSAR models, because every QSAR model has limitations and cannot be applied to all possible inputs. Conceptually, AD defines the region of ``normal'' objects in the chemical space, for which the QSAR model can be applied and get reliable predictions \cite{ROY2015231, mathea2016}. \cite{ROY2015231} categorizes different types of approaches for estimating AD, including ranges in the descriptor space, geometrical methods, distance-based methods, probability density distribution, and range of the response variable. 

Some approaches for AD estimation are widely used but do not apply to PADME, like the standard deviation of ensemble predictions, or the bounding box method, or in general descriptor space analysis \cite{sushko2010, ROY2015231}. In particular, descriptor space analysis is not applicable because the input of PADME does not involve chemical descriptors obtained through feature engineering, like Dragon descriptors. Furthermore, unlike typical QSAR models which only take compounds as input, PADME also has protein information as part of the input, adding a lot more complexity into AD estimation, like measuring the distance in the distance-based methods. Unlike conventional QSAR methods which only need to calculate the distance between compounds, for PADME, we have to consider the distance between compound-protein (drug-target) pairs, but to our knowledge, there is no widely accepted way of measuring the distance between those pairs. 

Thus, we decided to simply use the range of the response variable to define AD, which is a viable approach for regression models \cite{mathea2016, ROY2015231}. A natural method is to use mean and standard deviation, but since DTI datasets are often highly skewed (like the Davis dataset), using mean and standard deviation does not give reasonable AD range estimations. For example, in the 5-fold CV of Davis dataset, for one iteration, the training folds have a range \([5.0, 10.796]\), but the mean is 5.398, and the standard deviation is 0.8506. If we deem a test DT pair with response values lying within \((mean-k*std, mean + k*std)\) as being inside AD, where $k$ is some constant, $k$ must be very large to encompass the right end of the training set value range, but that would make the left end of the AD range too small to be realistic. 

\begin{figure}[ht] 

\includegraphics[width=\textwidth]{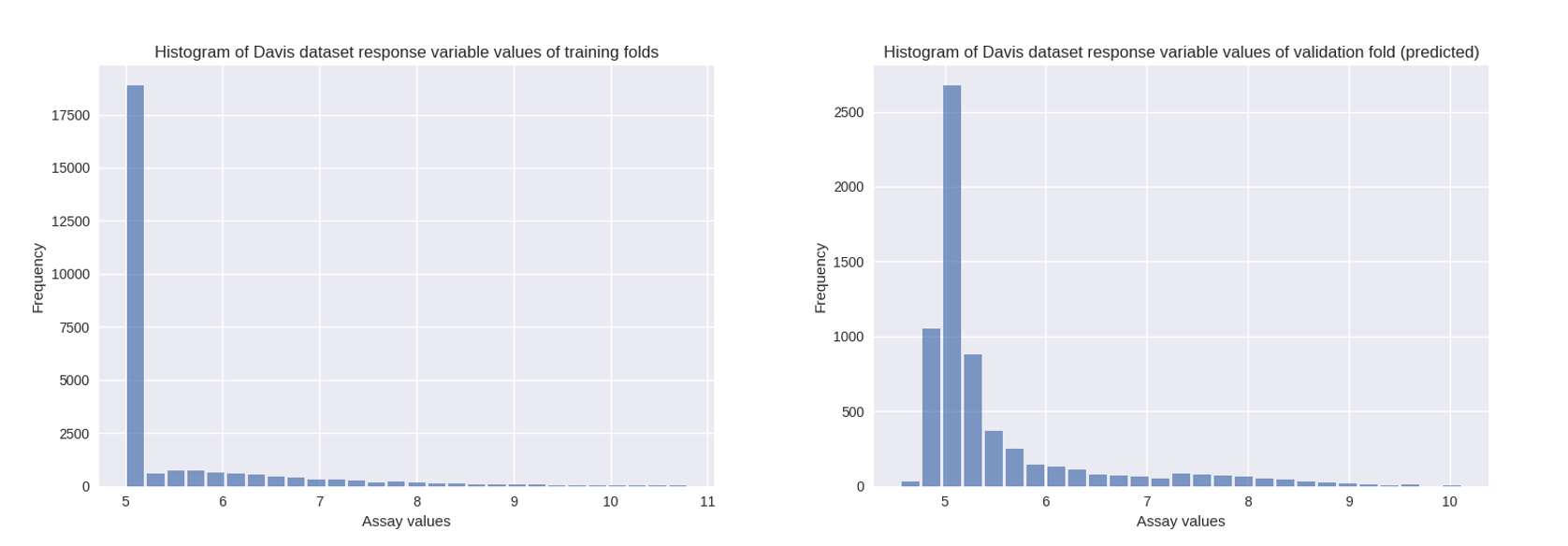}

\caption{The histograms of response variable values of the training folds and the predicted response variable values of the validation fold of a CV iteration of the Davis dataset. The range of response values of training folds is \([5.0, 10.796]\), while the range of predicted values in the validation fold is \([4.558, 10.123]\), demonstrating the validation fold elements are within the AD even if we don't know the true values of the response variable.}

\label{hist} 

\end{figure}

Instead, we propose the following simple method. Let \(min\) denote the minimum value of response variable $y$ in the training set, and \(max\) denote the maximum value in the training set, and the \(range size = max - min\). We define the AD to be \((min - 0.15*range size, max+0.15*range size)\). If a test drug-target pair has response value (experimental measurement) outside this range, we deem it as outside of AD. If the experimental measurement is unknown, we use the predicted response value as an approximation to the true response. \cref{hist} demonstrates that, in an iteration of CV of Davis dataset, the predicted response values in the validation fold lie within the aforementioned range calculated from the training folds. So we know the elements in the validation fold belong to AD, even if the true response values of the validation fold are unknown. Specifically, the range of response values of the training folds is \([5.0, 10.796]\), so the range size is 5.796. The AD is thus \([5 - 0.15*5.796, 10.796 + 0.15*5.796]\), which equals \([4.131, 11.665]\). The range of predicted values in the validation fold is \([4.558, 10.123]\), so all the predicted values lie in the AD range, thus all validation fold elements are in the AD.

\subsubsection{Qualitative Results}
\begin{figure}[!ht] 
\centering

\includegraphics[width=\textwidth, height=15cm, keepaspectratio]{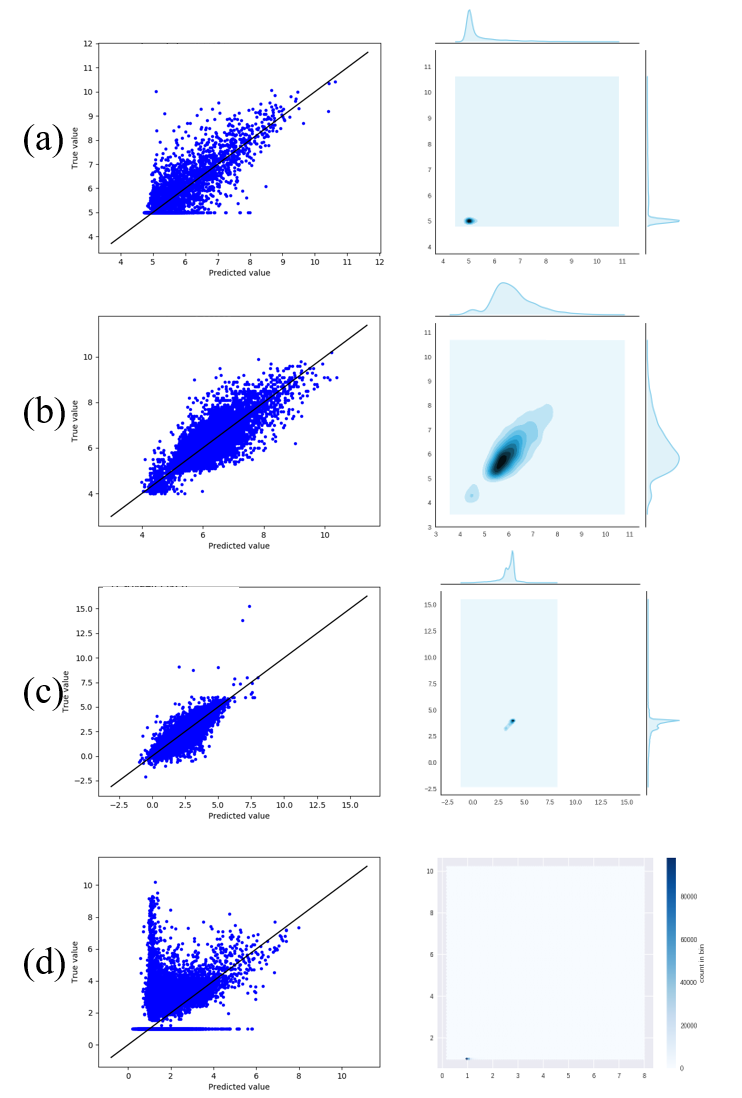}

\caption{\color{Gray}Scatter plot and contour plot of predicted VS true values across all datasets. The panels \textbf{a, b, c} and \textbf{d} correspond to \textbf{Davis, Metz, Kiba} and \textbf{ToxCast} datasets, respectively. The axes in the two plots of the same panel are the same, and both plots are generated from the same data. The diagonal lines in the scatter plots are the reference lines where \emph{predicted = true value}.}

\label{fig3} 

\end{figure}

\begin{figure}[!h] 


\includegraphics[width=5cm, keepaspectratio]{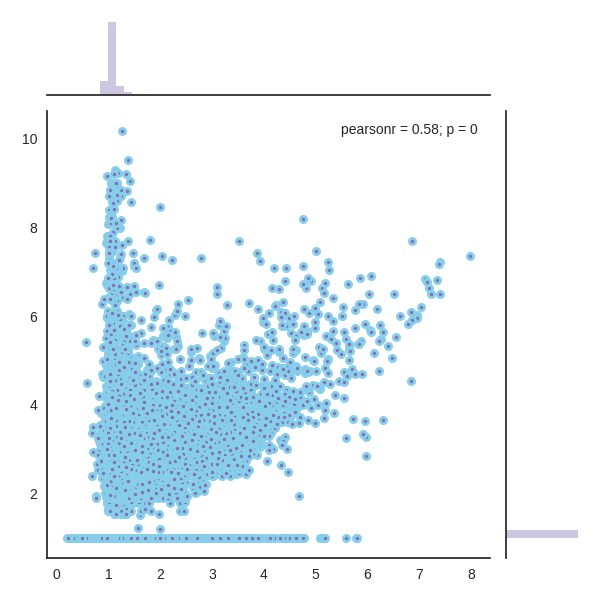}

\caption{ToxCast data scatter plot with marginal histograms, generated from the same data as Figure \ref{fig3}(d)}

\label{fig4} 

\end{figure}

We used plots to visualize the prediction performance, so we can assess the results qualitatively.

\cref{fig3} presents the predicted values (by PADME-ECFP) VS true values for each dataset. For each panel(row) in the figure, there is a scatter plot, and a contour plot (the darkness of the color represents the density of data points) with univariate density curves on the margins. Both plots in each row are plotted from the exact same data. Figure \ref{fig3}(d) is an exception, it includes a hexagon plot instead of a contour plot, because the contour plot fails to show anything (but the density curves on the margins are plotted), possibly because the data points of ToxCast are too concentrated to be shown correctly on the contour plot, as can be observed from the hexagon plot. To help visualize ToxCast better, we added a Figure \ref{fig4} which is a scatter plot of the same data 
as Figure \ref{fig3}(d), with histograms on the margins. 

Clearly, all datasets except Metz data are very concentrated at some values.

\begin{figure}[!ht] 


\includegraphics[width=\textwidth]{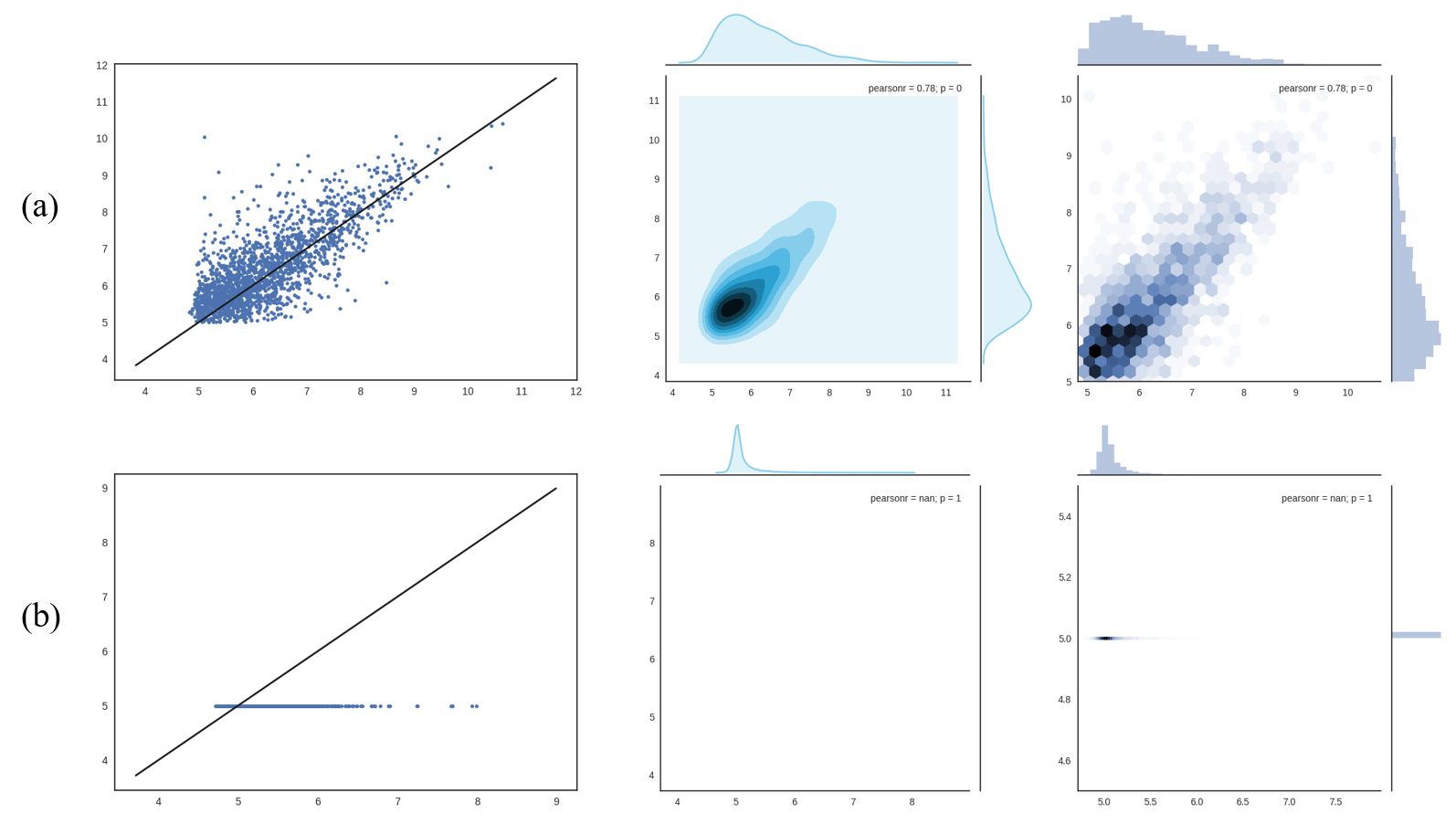}

\caption{\color{Gray}Plots for Davis dataset predicted value VS true value. Panel (a) corresponds to the true active values, while panel (b) corresponds to true inactive values. Similar to figure \ref{fig3}, all plots in the same panel are plotted from the same data.}

\label{fig5} 

\end{figure}

\begin{figure}[!ht] 


\includegraphics[width=\textwidth]{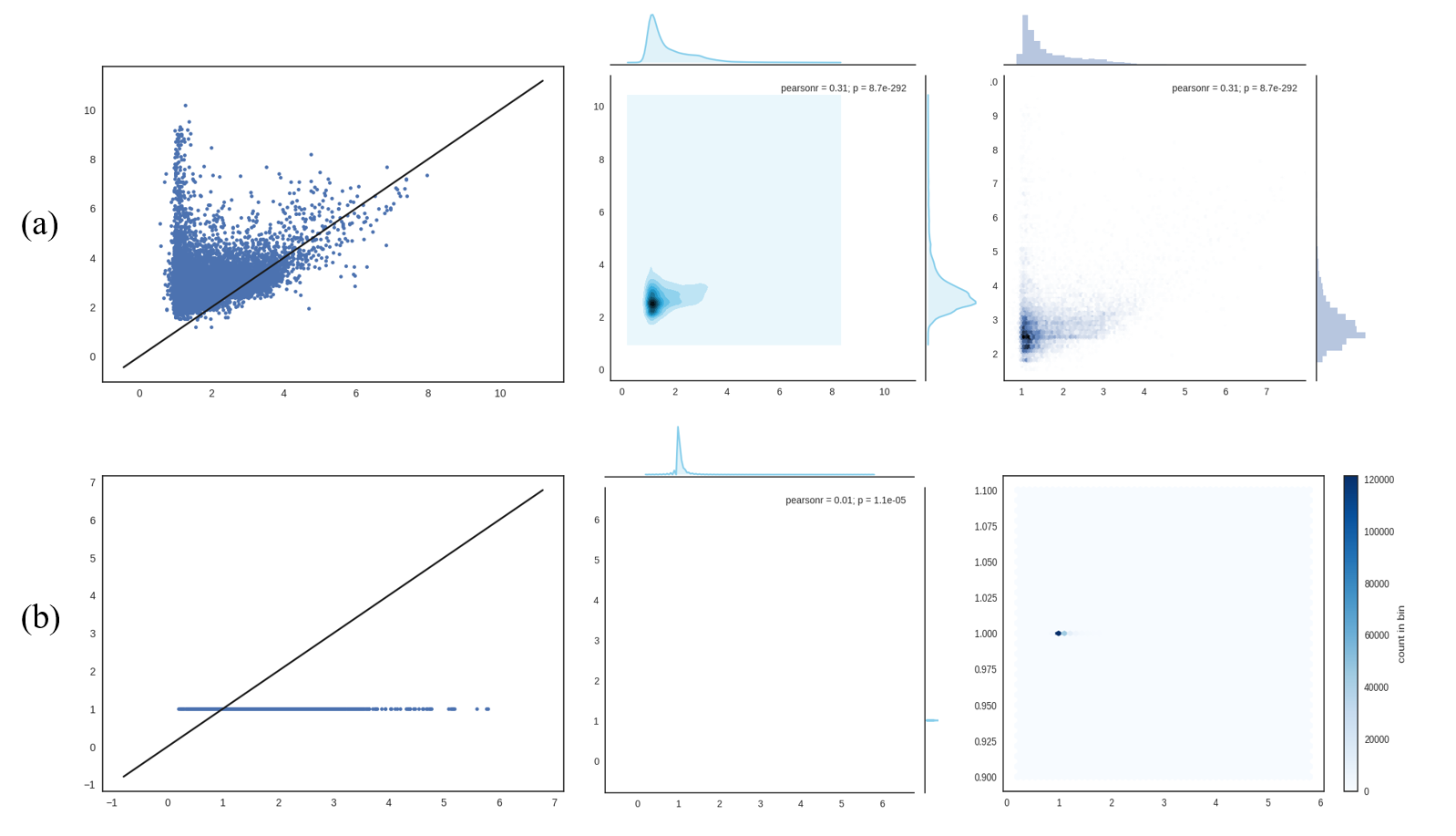}

\caption{\color{Gray}Similar to Figure \ref{fig5}, plots for the ToxCast dataset. Panel (b) uses a different hexagon plot from (a), because that form of hexagon plot on panel (b) does not display properly.}

\label{fig6} 

\end{figure}

Because the concentration of Davis and ToxCast datasets pose problems in visualizing the prediction performances on them, we decided to plot the scatter plots, contour plots and hexagon plots of the true active and true inactive data points separately for those datasets (\cref{fig5,fig6}). From \cref{fig5} we can see the Davis dataset was predicted pretty well on both true active and inactive values, \cref{fig5} (a) shows a nice pattern of correspondence between predicted VS true values on the active data points, while \cref{fig5} (b) presents true inactive values, in which the hexagon plot shows a high concentration of predicted values close to the true values. As reflected in \cref{fig6} (a), the model fitted on the ToxCast dataset was strongly influenced by inactive values, and the prediction performance for the true active data points was not very good, but the true inactive data points were predicted to concentrate around the true values (shown in the hexagon plot in \cref{fig6} (b)), which might explain why its quantitative analysis results were decent.

To tackle the imbalanced dataset problem in ToxCast, we tried to train a model on the oversampled dataset and measured its performances. Please refer to the Supporting Information.

So why does PADME perform well on Davis, Metz and KIBA datasets, but not so satisfactorily on the ToxCast dataset? We think it might be related to the nature of the ToxCast dataset itself. The ToxCast dataset not only contains a much larger variety of proteins (unlike the other 3 datasets which only contain kinase inhibitors), but it also has a much larger number of assays (measurement endpoints), which are often quite different from each other. Though we only selected the assays with single intended targets, many of those assays are cell-based (for example, OT\_AR\_ARSRC1\_0480), which could introduce some more complexities in addition to the drug-target interaction, due to the intricacies of biochemical processes in cells. These challenges might be some reasons why, to our knowledge, previous non-docking researches on drug-target interaction prediction containing protein information as input did not use this dataset \cite{He2017,Pahikkala2014,wen2017,ozturk2018}, though other kinds of researches did \cite{chushak2018,mansouri2016,liu2015}.

The challenges with the ToxCast dataset, including its large number of measurement endpoints and the imbalanced dataset problem, should be investigated in future work, as it is an important objective to build a more general-purpose DTI prediction model that handles a larger variety of input proteins, compounds and measurement endpoints. Imbalanced datasets also frequently appear in virtual screening. We believe that, based on our work, future researches will either make improvements on the ToxCast dataset or deem it as a great challenge for DTI prediction models involving protein information, and our results presented here, though not ideal, is of reference value to the community.

\subsubsection{Case Study}
In addition to the quantitative and qualitative evaluations shown above, we performed a case study to further validate the predictions of PADME by investigating the compounds predicted to interact strongly with selected target proteins.

We focused on the androgen receptor (AR), for which alterations of functions are associated with prostate \cite{yap2016} and breast cancers \cite{mina2017}. 

We used all compounds in the datasets used in this paper, together with all the compounds in US National Cancer Institute human tumour cell line anticancer drug screen data (NCI60), totaling more than 100000, and AR as the only target protein. NCI60 dataset records the \emph{in vitro} drug response of cancer cell lines \cite{shoemaker2006}. For prediction, we used PADME-ECFP and PADME-GraphConv trained on the whole ToxCast dataset, then took the average of their predictions, we call this averaged model PADME-Ensemble. The reason we chose the ToxCast dataset is that its endpoints are the most suitable for calculating AR binding affinity or antagonistic effects, and the ToxCast dataset has the most diverse set of compounds and proteins. 

There are many different assays in ToxCast, some are cell-based, while some are cell-free. Cell-based assays are much more complicated than their cell-free counterparts, since the results of cell-based assays might involve some intricate biochemical reactions in the cells. Thus, we used the assay NVS\_NR\_hAR, a cell-free assay measuring the binding affinity between Human Androgen Receptor (AR) and ligands (please refer to the Supporting Information for details), to examine the efficacy of PADME's predictions. 

From the predictions of PADME-Ensemble, we selected the prediction results corresponding to NVS\_NR\_hAR, and then sorted the predicted values in descending order. Due to the transformations we performed in \autoref{subsec: datasets}, the larger the number (higher in the sorted list), the stronger the binding affinity. Next, we filtered out those compounds in the predicted list that have appeared in the ToxCast dataset, or had a large Tanimoto similarity with some compounds in ToxCast, calculated using rdkit fingerprint. We then did a search on PubChem database \cite{kim2018} for the top 30 compounds predicted to bind strongly with AR.

The top 30 (and beyond) compounds all shared a highly similar structure with androgen, so most of them may be able to bind strongly with AR. After a stringent search on PubChem, we confirmed that 4 of them are active, which is reflected in patents, bioassay results, or research papers. The other compounds in top 30 are also possibly active, but since there is no direct evidence from PubChem, we take the conservative approach and do not consider them here.

\begin{figure}[!h] 

\includegraphics[width=12cm, keepaspectratio]{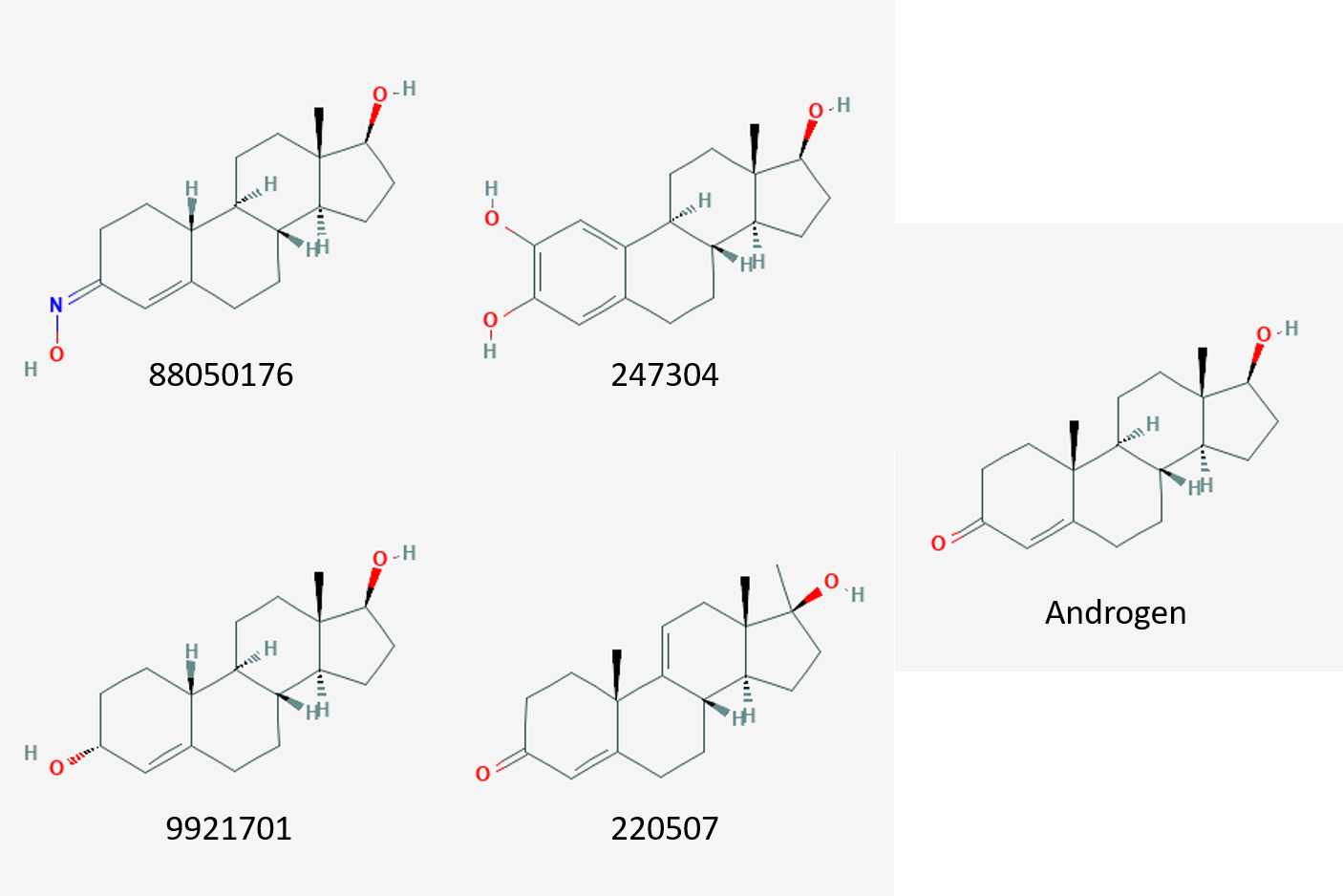}

\caption{The 4 compounds from top 30 predictions that are confirmed to bind strongly with AR. The numbers are their corresponding PubChem CIDs. On the right side is the 2d representation of testosterone, the major androgen. The images are downloaded from the PubChem website.}

\label{5_cpds} 

\end{figure}

The 4 compounds' PubChem CID are 88050176, 247304, 9921701, 220507. \cref{5_cpds} shows their 2D images.

Obviously, the compounds are all very similar to androgen, since NVS\_NR\_hAR is a very simple assay, the model learns from the dataset that analogs of androgen tend to bind strongly with AR. This shows that the prediction results of PADME are effective in drug discovery.

Based on this, we tried to take one step further to do a more interesting task: calculating the AR antagonist effect of compounds based on the predictions produced by PADME. Because there are 61 outputs in PADME models trained on ToxCast data, we had to propose a set of coefficients to calculate a composite AR antagonist score (details in the Supporting Information) from the averaged prediction results, for which we expect the compounds with higher scores would show stronger anticancer activity in AR-related cell lines in NCI60 dataset. We then ordered the AR antagonist scores in descending order.

Compared to those predicted to bind strongly with AR, the predicted list of AR antagonistic compounds have much more diverse structures. However, their results (see Supporting Information) are not well aligned with our expectations. This is not caused by PADME, which captures the patterns in the training data faithfully and shows it in the test set. It is the results on the true dataset that are different from our assumptions.

There are two major challenges in this process: the formula we used for calculating AR antagonist score is based on assumptions, the existence of cell-based assays is also a possible source of problem; our expectation that AR antagonistic compounds should perform selectively on some known AR-related cancer cell lines might deviate from the truth, or AR influences many types of cancers in different ways from what we knew, like suggested in  \cite{munoz2015}. Tackling these two challenges is a task that could require years or decades of work by the community. 

Although the AR antagonist score is problematic, we still decided to check whether it is useful. We chose 38 compounds from the top predictions for AR antagonists that were available from the vendors including the National Cancer Institute \cite{NCI2019}, and 3 of them were confirmed to be active, which have ZINC IDs ZINC8665890, ZINC3861637, and ZINC4947964. Please see Supporting Information for details. 

In all, the obtained results indicate that PADME is capable of identifying compounds that have the desired simple interactions with the target protein.


\section{Discussion}
PADME is compatible with a large variety of different protein and compound featurization methods. By combining different protein descriptors like PSSM and other molecule featurization schemes like Weave \cite{kearnes2016}, many more variants of PADME can be constructed, whose performances can be compared against each other. In fact, we used the Weave implementation in DeepChem as a molecular featurization method and ran hyperparameter searching on it, but the result was worse than ECFP and GraphConv, and it consumed much more time and memory than the other two, so we did not pursue it any further. 

Mapping a protein into a feature vector is a task in proteochemometrics. However, most existing methods in proteochemometrics require expert knowledge and often involve 3D structural information \cite{van2011, qiu2016}, which is often not available. We only considered sequence information for both drugs and targets in this work to make our model more generally applicable.

It is also possible to use CNN or RNN to learn a latent feature vector to represent the proteins, based on its amino acid sequence information, instead of using fixed-rule protein descriptors like PSC as the input, so that the whole model can be trained in a completely end-to-end fashion without standalone components of the network like PSC in our implementation, making the network structure more ``symmetrical''. Actually, it was already attempted by {\"O}zt{\"u}rk et al. \cite{ozturk2018}, who showed a performance similar to PADME, but they did not use cross-validation to get average performances, they only ran different models on the same test set, which was just 1/6 the size of the whole dataset, so we think there is still much room for improvement in that direction. Nonetheless, it is a very good step towards it.

We only used simple feedforward neural networks in our implementations of PADME from the Combined Input Vector to the output, but other types of Neural Networks might be able to generate better results, like Highway Networks \cite{srivastava2015}, which allows the units in the network to take shortcuts, circumventing the large amounts of layers in some networks. 

Pretraining also has the potential to improve our model, but we did not include it, because it might be difficult for the community to compare the real performance of PADME with other models.

Compared to previous models like SimBoost and KronRLS, PADME is not only outperforming them in terms of prediction accuracy, but is more scalable in terms of number of drugs and targets and number of prediction endpoints\footnote{As elaborated in \autoref{subsec: tc}, the scalability in the number of drugs and targets might not always be the case, but the scalability in the number of prediction endpoints is.}, because both SimBoost and KronRLS rely on similarity matrices and they are only single-task models. In the age of Big Data, this scalability will be a big advantage in virtual high-throughput screening.

\section{Conclusion}
To tackle DTI regression problem more effectively, we devised the PADME framework that utilizes deep neural networks for this task. PADME incorporates both compound and target protein sequence information, so it can handle the cold-start problem, which most current deep learning-based models for DTI prediction cannot do. Using sequence information as the input makes the model simple and generally applicable. Predicting real-valued endpoints also makes it desirable for problems requiring finer granularity than binary classification.

PADME is the first method to incorporate MGC with protein descriptors into the DTI prediction task, and has been shown be consistently outperform state-of-the-art methods as well as Compound-Only DNN models. Surprisingly enough, PADME based on MGC (GraphConv in our case) does not outperform PADME based on ECFP, which could be due to the difficulty of finding the best set of hyperparameters for MGC. More work is needed to construct a better MGC model or find a better set of hyperparameters for the existing model. PADME is also more scalable than the state-of-the art models for DTI regression task, namely SimBoost and KronRLS, and this advantage might be significant in datasets with lots of compounds/targets and multiple measurement endpoints. Another contribution is the use of the ToxCast dataset in DTI prediction problems with protein information input, which we believe future research should investigate further in addition to the other benchmarking datasets. Our results on the ToxCast dataset suggests it is a greater challenge than we expected.

As a case study, we predicted the binding affinity between compounds and the androgen receptor (AR), a high proportion of the compounds predicted to bind strongly with AR are confirmed through database/literature search. This suggests that PADME has the potential to be applied in drug development, and will likely benefit domains like toxicity prediction, computer-aided drug discovery, precision medicine, etc. 

With the compatibility of PADME to different drug molecule and target protein featurization methods, as well as its scalability compared to methods which rely on similarity matrices and have single outputs, we believe that future work could propose more PADME variants that advance the frontier of DTI prediction research.


\section*{Supporting Information}
The following files are available free of charge.
\begin{itemize}
  \item PADME\_supplementary\_dataset\_files.zip: will be available in journal version of this paper. Exceeds size limit of arXiv.
  \item supplementary\_text.pdf: explains data preprocessing steps and presents some additional experiments. 
\end{itemize}

The source code and some processed datasets are deposited at https://github.com/simonfqy/PADME. Some bigger processed datasets could be obtained upon request.


\section*{Acknowledgments}
The authors thank Dr. Fuqiang Ban and Dr. Michael Hsing for giving us some useful information that we incorporated into this paper. Helene Morin and Eric LeBlanc were in charge of wet-lab experiments for validating our results. We also thank the help and suggestions received from other fellow lab members, including but not limited to Zaccary Alperstein, Oliver Snow, Michael Lllamosa, Hossein Sharifi, Beidou Wang, Jiaxi Tang and Sahand Khakabimamaghani. We also express our gratitude towards our family and friends, especially Jun Li, Wen Xie, Qiao He, Yue Long, Aster Li, Lan Lin, Xuyan Qian, Zhilin Zhang, Qing Rong, Si Chen, Fengjie Lun and Stephen Tseng, the list goes on; most important of all, the late Mr. Xiefu Zang.

We also thank George Lucas (and his prequel trilogy) and Smith et al. \cite{smith2017} for the inspiration for naming.

\bibliography{library}
\bibliographystyle{abbrv}
\end{CJK*}
\end{document}